\documentclass[lettersize,journal]{IEEEtran}
\usepackage{amsmath,amsfonts,amssymb}
\usepackage{algorithmic}
\usepackage{algorithm}
\usepackage{array}
\usepackage[caption=false,font=normalsize,labelfont=sf,textfont=sf]{subfig}
\usepackage{textcomp}
\usepackage{stfloats}
\usepackage{url}
\usepackage{verbatim}
\usepackage{graphicx}
\usepackage{cite}
\usepackage{booktabs}
\usepackage{multirow}
\newcolumntype{L}[1]{>{\raggedright\arraybackslash}p{#1}}
\newcolumntype{C}[1]{>{\centering\arraybackslash}p{#1}}
\usepackage{graphicx}
\usepackage[hidelinks]{hyperref}
\usepackage{orcidlink}
\begin{document}

\title{TACTUS: A Robotic Palpation System for Spatially Heterogeneous Hardness Mapping of Woody Breast Myopathy}

\author{Chaitanya~Pallerla\orcidlink{0000-0003-1211-9036}$^{1,2}$,
Yihong~Feng\orcidlink{0009-0006-4220-6607}$^{1}$,
Avisekh~Tripathy\orcidlink{0009-0005-7415-0495}$^{1,2}$,
Yang~Tian\orcidlink{0009-0004-9347-1342}$^{1}$,
Pouya~Sohrabipour\orcidlink{0009-0009-4068-4818}$^{1}$,
Amirreza~Davar\orcidlink{0009-0003-2893-8746}$^{1}$,
Casey~M.~Owens\orcidlink{0000-0003-1382-6904}$^{3}$,
Cengiz~Koparan\orcidlink{0000-0003-3940-0055}$^{1}$,
and~Dongyi~Wang\orcidlink{0000-0002-1224-5529}$^{1,2,*}$
\thanks{$^{1}$Department of Biological and Agricultural Engineering, University of Arkansas, Fayetteville, AR, USA, 72701.}
\thanks{$^{2}$Department of Food Science, University of Arkansas, Fayetteville, AR, USA, 72701.}
\thanks{$^{3}$Department of Poultry Science, University of Arkansas, Fayetteville, AR, USA, 72701.}
\thanks{*Corresponding author: Dongyi Wang (e-mail: dongyiw@uark.edu).}
}

\markboth{}%
{Pallerla \MakeLowercase{\textit{et al.}}: TACTUS: A Robotic Palpation System for Spatially Heterogeneous Hardness Mapping of Woody Breast Myopathy}

\maketitle

\begin{abstract}
Woody breast (WB) myopathy is a spatially heterogeneous fibro-degenerative condition of the broiler \textit{Pectoralis major} muscle, yet existing mechanical reference methods rely on single-point or sequential manual compression measurements that cannot capture this heterogeneity, while non-invasive imaging systems lack spatially resolved ground-truth data. This paper presents TACTUS (Tactile Adaptive Conformal three dimension (3D) Uniform Sensing), a tactile array that performs automated, thickness-normalized, 3D hardness mapping of broiler breast fillets at 20 anatomically distributed locations in a single \textasciitilde30~s acquisition cycle. TACTUS integrates 20 independently motorized tactuators, each comprising a force-sensing resistor and quadrature-encoder-driven linear actuator, with a LiDAR depth camera for pre-compression surface profiling, enabling conformal, thickness-normalized compression across irregular fillet geometry. The system was evaluated on 100 broiler fillets spanning four woody breast severity classes (Normal, Mild, Moderate, Severe). TACTUS detected significant force differences across severity categories (ANOVA, $p<0.001$), with the strongest differentiation in the cranial region and a significant caudal-region effect not captured by texture-analyzer measurement. Fillet-level TACTUS force showed moderate but significant agreement with texture-analyzer measurements (Pearson $r = 0.441$, $p<0.001$). Semivariogram, local change entropy, and local heterogeneity mapping analyses revealed that hardness variability increases with WB severity and follows a spatially structured, anatomically localized pattern rather than a uniform distribution, quantifying aspects of WB pathology invisible to single-point measurement. These results establish TACTUS as an automated, operator-independent platform for generating spatially registered mechanical hardness maps, offering a scalable ground-truth tool for training and benchmarking hyperspectral and near-infrared spectroscopy-based WB classification systems.
\end{abstract}


\section{Introduction}
\IEEEPARstart{W}{oody} breast (WB) myopathy has become one of the most economically consequential muscle quality defects in the commercial broiler industry worldwide. First formally characterized by Sihvo et al. \cite{ref8}, the condition manifests as multifocal myodegeneration accompanied by marked fibrosis, inflammatory infiltration, aberrant collagen cross-linking, and disrupted calcium homeostasis within the \textit{Pectoralis major} muscle of rapidly growing broiler strains. Prevalence estimates across commercial broiler flocks range from 10\% to 55\% for moderate to severe woody breast, whereas total woody breast prevalence has been reported as high as 82.4\%, with 11.9\% of fillets classified as severe at slaughter, depending on genetic line, slaughter age, production conditions, and severity classification \cite{ref1,ref2}. Annual economic losses in the United States alone are estimated to exceed USD 200 million due to downgrading, consumer rejection, and impaired processing yield \cite{ref3}. Beyond its economic impact, WB has significant implications for meat safety and consumer acceptance: affected fillets exhibit altered water-holding capacity, elevated pH, and compromised protein functionality, rendering them unsuitable for premium marinated and further-processed product lines \cite{ref4,ref5}.

The defining biomechanical hallmark of WB is tissue hardness heterogeneity. Unlike uniform myopathies that affect the whole muscle symmetrically, WB produces a spatially complex hardness landscape in which localized fibrous nodules and hardened plaques co-exist with regions of near-normal soft tissue within the same fillet \cite{ref6}. This heterogeneity follows anatomically structured patterns, with the cranial lobe and lateral margins disproportionately affected in moderate and severe cases, while mild WB frequently manifests as small, internally embedded stiff foci that produce minimal surface optical contrast and no visible external deformation \cite{ref7}. The spatial distribution and extent of these mechanically stiff regions reflect the underlying histopathological changes associated with woody breast, including muscle fiber necrosis, collagen deposition, and regenerative activity. Consequently, spatially resolved hardness mapping provides an objective means of characterizing both the severity and anatomical distribution of woody breast lesions \cite{ref8,ref9}.

Despite this well-established spatial nature of the defect, the methodological toolkit for WB characterization remains fundamentally point-localized. The most widely adopted mechanical reference methods are compression force (CF) measured by a texture analyzer probe, the Meullenet-Owens Razor Shear (MORS) \cite{ref10}, and its blunt variant (BMORS). These methods capture the mechanical response at a single predetermined location, typically the anatomically thickest region of the cranial lobe \cite{ref11,ref12}. While CF shows moderate to strong correlations with subjective WB scores, this localized measure fails entirely to represent the spatial extent of fibrosis or the degree of hardness heterogeneity within the fillet. Fillets with comparable mean hardness can exhibit dramatically different spatial distributions, one showing a single large central plaque, another showing diffuse peripheral involvement, and these differences have direct implications for processing behavior, consumer texture perception, and the training of imaging classifiers \cite{ref6}.

The critical importance of spatial hardness mapping was demonstrated by Parajuli et al. \cite{ref6}, who applied a systematic grid-based CF protocol to map hardness variation across the fillet surface, revealing spatial heterogeneity of hardness distribution that correlated with histological fibrosis scores. This study provided strong conceptual proof that WB heterogeneity can be meaningfully quantified through spatially distributed force measurements. However, the protocol required sequential manual probe repositioning at 20--30 individual grid locations per fillet, was inherently operator-dependent, could not guarantee perpendicular probe contact on curved surfaces, and required approximately 15--20 minutes per fillet, making it entirely impractical for studies requiring large sample sizes or for any form of inline industrial monitoring.

Concurrently, a substantial body of literature has applied non-invasive sensing technologies to WB detection with the goal of enabling high-throughput, non-destructive classification. Hyperspectral imaging (HSI) in the visible and near-infrared range has achieved promising classification accuracies for moderate and severe WB by exploiting spectral features associated with water binding capacity status, collagen fluorescence, and heme absorption (\cite{ref13,ref7,ref14}). Near-infrared spectroscopy (NIRS) has been applied for rapid in-line estimation of WB severity, with partial least-squares regression models explaining up to 70\% of variance in subjective scores \cite{ref15}. Thermal imaging has been explored as a proxy for metabolic activity differences between normal and fibrotic tissue \cite{ref16}, while ultrasound and optical coherence tomography (OCT) have demonstrated sensitivity to subsurface structural changes associated with early fibrosis \cite{ref7}.

Despite this diversification of non-invasive approaches, a fundamental methodological gap has persisted: none of these imaging modalities can directly measure tissue hardness, and the ground-truth labels used to train and evaluate their classification models are almost universally based on subjective palpation scores or single-point CF measurements. The consequence is a systematic mismatch between what non-invasive methods detect, spatially varying optical, spectral, or thermal signals distributed across the fillet surface, and the scalar, location-specific labels used for model training. A classifier trained on cranial-lobe CF values, for instance, learns to predict central fillet hardness but is never informed about the spatial distribution of hardness that its spectral features are actually encoding. This mismatch constrains achievable classification accuracy and prevents the extraction of spatially rich predictive features from imaging data.

The broader field of robotic tactile sensing offers engineering precedents for distributed force measurement, but existing systems are not suitable for the WB application without substantial re-engineering. Vision-based tactile sensors such as GelSight \cite{ref17} and GelSlim \cite{ref18} provide extremely high spatial resolution within a small contact patch but are fundamentally single-contact devices that cannot simultaneously sample distributed locations across a 100~$\times$~125~mm surface. Flexible and camera-based tactile sensors have also been applied to non-destructive food firmness testing, including fruit firmness assessment \cite{ref19} and tactile contact-shape recognition for soft-tissue interaction \cite{ref20}, achieving only localized or single-contact measurement. However, these systems employ rigid or semi-rigid substrates that cannot conform to the complex dorsal curvature of a boneless chicken fillet, leading to non-uniform contact pressure, missed contact in recessed zones, and systematic force measurement error. Shape-adaptive soft robotic end-effectors \cite{ref21,ref22} achieve conformability through passive deformation but are not instrumented for distributed force sensing. No existing system integrates motorized conformal actuation, simultaneous distributed force sensing, LiDAR-based thickness measurement for geometric normalization, and robotic positioning within a single platform.

This paper introduces TACTUS (Tactile Adaptive Conformal 3D Uniform Sensing), the first purpose-engineered robotic tactile array for three-dimensional hardness mapping of WB fillets. The key innovations of TACTUS are fourfold. First, independent motorized actuation of each tactuator enables the array to conform to arbitrary fillet surface geometry, ensuring full simultaneous contact at all 20 measurement points regardless of surface curvature during a single measurement cycle. Second, LiDAR-based pre-compression depth profiling enables local fillet thickness estimation at each measurement point, supporting compression depth normalization that produces thickness-independent hardness values comparable across fillets of different sizes. Without thickness normalization, thicker regions or fillets may exhibit higher compression forces solely due to increased tissue thickness rather than greater mechanical stiffness, confounding comparisons across locations and samples. Third, the 4$\times$5 spatial array covers the full dorsal fillet surface in a single acquisition cycle, generating a spatially registered hardness map that enables the computation of a novel spatial statistics heterogeneity index, hardness entropy, and Moran's I with direct biological relevance to WB fibrosis patterns. Fourth, UR5e robotic deployment provides controlled, repeatable, operator-independent positioning and force application, eliminating the principal sources of inter-measurement variability that afflict manual CF protocols. Together, these capabilities satisfy the four engineering requirements necessary for high-resolution spatial mechanical characterization of heterogeneous materials.

A comparative analysis of representative tactile sensing systems developed for food, agricultural, and biomedical applications is presented in Table~\ref{tab:comparative}. The comparison demonstrates that no existing platform combines these four capabilities for quantitative three-dimensional mechanical mapping. Although the present study validates the system using woody breast myopathy as a representative case study, the proposed TACTUS platform is broadly applicable to heterogeneous materials exhibiting spatially varying mechanical properties. Beyond WB assessment, the modular design of TACTUS provides potential for broader mechanical characterization of food and biological materials. The system could be adapted to map spatial variations in tenderness, firmness, or other mechanical properties in beef, pork, lamb, fish, other poultry cuts, and whole-carcass regions; characterize firmness changes in cheese during ripening; assess maturity or mechanical defects in horticultural products; evaluate dough and other processed foods; and characterize localized mechanical changes in soft biological tissues. Because TACTUS can operate with partial sensor coverage, samples do not need to match the full 4~$\times$~5 array footprint. Larger or irregularly shaped samples can therefore be evaluated without hardware modification, although the fixed array size and spacing would provide proportionally lower coverage of the complete sample surface. Importantly, the 20 tactuators can be controlled independently, allowing users to activate selected tactuators and perform targeted measurements at specific regions of interest rather than compressing the entire array. This feature also enables customized compression patterns and potentially different measurement protocols across regions of the same sample.

\begin{table*}[!t]
\caption{Comparative Survey of Tactile Sensing and Distributed Force Measurement Systems Across Food, Agricultural, and Biomedical Applications, Illustrating the Capability Gap Addressed by TACTUS}
\label{tab:comparative}
\centering
\resizebox{\textwidth}{!}{%
\begin{tabular}{@{}L{2.6cm}L{1.8cm}L{2.6cm}C{1.5cm}C{1.5cm}L{2.4cm}L{4.3cm}@{}}
\toprule
\textbf{System/Study} & \textbf{Domain} & \textbf{Sensing Modality} & \textbf{Array Size} & \textbf{Spatial Map/Depth Sensing} & \textbf{Key Application} & \textbf{Limitation vs. TACTUS} \\
\midrule
GelSight \cite{ref17} & Robotics & 2D vision-based gel contact & Single contact & Y/N & Surface texture and force & Dense optical tactile imaging without independent force sensing or thickness normalization. \\
BioTac \cite{ref23} & Surgical robotics & Fluid pressure + vibration & Single contact & N/N & Fingertip haptics & Single point only; no spatial map \\
Parajuli et al. \cite{ref6} & Food (poultry) & Manual CF probe (sequential) & Manual grid & Y/N & WB hardness mapping & Slow; manual; operator-dependent; no automation \\
Ma et al. \cite{ref19} & Agriculture (fruit) & Visuo-tactile (camera + marker) & Single sensor & Within patch & Peach firmness & Single contact; no distributed array; not conformal to fillet \\
Liu et al. \cite{ref20} & Medical & Flexible piezoresistive array for 2D & 8$\times$8 & Y/N & Soft tissue palpation & No thickness normalization; no depth integration \\
Conde et al. \cite{ref24} & Food quality & Impact force-deformation & Single point & N/N & Cheese firmness & Single point; not spatially resolved; no depth normalization \\
Narang et al. \cite{ref25} & Soft robotics & Distributed tactile skin & Many taxels & Y/N & Object classification & Not food-specific; no LiDAR depth; no normalization \\
\textbf{TACTUS (this study)} & \textbf{Food (poultry)} & \textbf{Thin-film FSR array + LiDAR} & \textbf{4$\times$5 (20 pts)} & \textbf{Y/Y} & \textbf{3D spatial hardness} & \textbf{---} \\
\bottomrule
\end{tabular}%
}
\end{table*}

The primary objective of this work is to describe the engineering design, operating protocol, calibration methodology, and validation framework of TACTUS, and to establish its utility as a high-fidelity mechanical ground-truth generation platform. The 3D hardness maps produced by TACTUS are intended to replace single-point CF measurements as the reference standard for WB severity grading, enabling both a more accurate characterization of fillet heterogeneity and the generation of spatially registered label datasets for training and benchmarking of imaging-based WB classification systems operating across the full fillet surface. The following manuscript is organized as follows. Section~II describes the materials and methods, including sample preparation, the TACTUS system architecture, robotic operating protocol, LiDAR-based thickness normalization, hardness map generation, sensor calibration, and statistical analysis. Section~III presents the results and discussion, including thickness-normalized hardness measurement, spatial mechanical phenotyping of woody breast, anatomical distribution of mechanical alterations, and characterization of spatial fibrosis patterns. Finally, the manuscript concludes with a discussion of the study limitations, future research directions, and the overall conclusions.

\section{Materials and Methods}

\subsection{Sample Preparation}
This study was conducted at the University of Arkansas's Biological and Agricultural Engineering laboratory. A total of 100 Cobb broilers of 8 weeks of age were randomly selected from the University's poultry farm and processed in the University of Arkansas pilot processing plant. Broilers were processed using commercial methods, and fillets were deboned from carcasses approximately 30~min after postmortem. Then, 100 chilled chicken breast samples, including the left and right sides, were collected and scored in the processing plant. These samples were classified as normal breast (NB), mild woody breast (MWB), moderate woody breast (MoWB) and severe woody breast (SWB) using hand palpation by a specialist (NB~=~20; MWB~=~35; MoWB~=~20; SWB~=~25). After WB scoring, these samples were moved to the laboratory for data collection. This processing protocol followed the procedure for sample preparation on the previous study \cite{ref1}.

\subsection{Data Collection}
For data collection, the mechanical hardness of the raw Woody Breast fillets was first measured using a motorized texture analyzer (MultiTest-dV, Mecmesin, Slinfold, United Kingdom) operated using VectorPro software (version 6.16.0.0). Hardness was quantified as the compression force (N) required to compress the tissue to 20\% of the local fillet thickness at each measurement location. A 1~$\times$~1~cm flat probe was used to apply perpendicular compression to the fillet surface without puncturing the tissue. Each breast fillet was divided anatomically into three regions, cranial, medial, and caudal, and three measurement points were collected from each region, resulting in nine texture-analyzer measurements per fillet. All nine measurements were retained for subsequent analysis. Following texture-analyzer measurements, the same fillets were evaluated using the TACTUS robotic tactile array. TACTUS applied the same thickness-normalized compression principle, with each active tactuator compressing the tissue to 20\% of the local fillet thickness. Because the number of tactuators contacting the fillet depended on its size, shape, and position beneath the array, approximately seven active measurement points were obtained per compression cycle on average. Each fillet was subjected to two TACTUS compression cycles. Between cycles, the fillet was repositioned so that the second compression was applied at different surface locations, thereby avoiding repeated compression of the same tissue points and increasing spatial coverage. The resulting TACTUS force measurements were retained for subsequent spatial hardness analyses and comparison with the texture-analyzer measurements.

\begin{figure*}[!t]
\centering
\includegraphics[width=\textwidth]{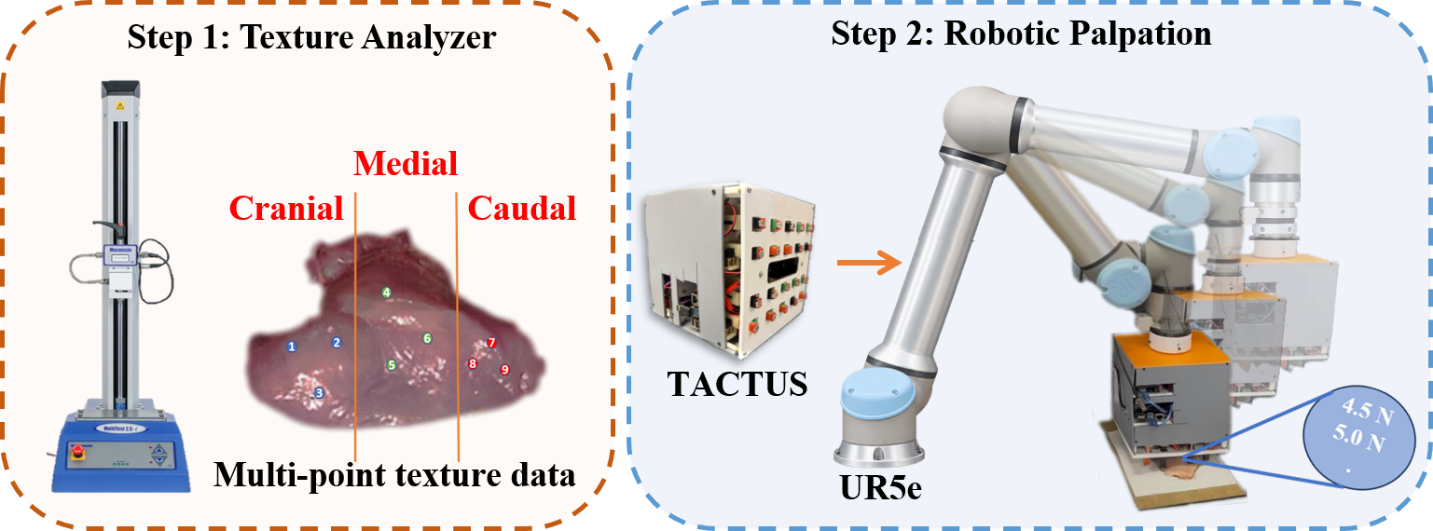}
\caption{The diagram of TACTUS based data collection workflow.}
\label{fig:1}
\end{figure*}

\subsection{TACTUS Device Architecture and Mechanism}
\subsubsection{Overall Hardware Architecture}
\begin{figure*}[!t]
\centering
\includegraphics[width=\textwidth]{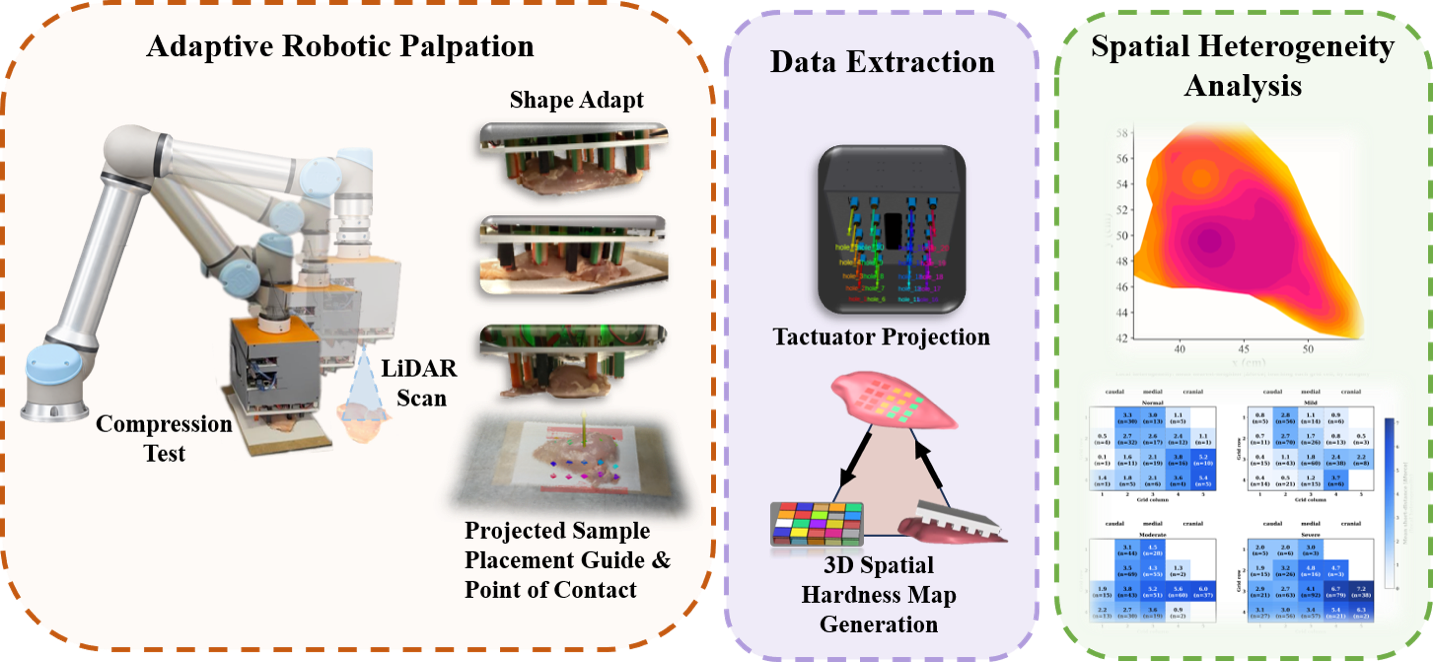}
\caption{Workflow of TACTUS operations. The UR5e arm carries the TACTUS device over the fillet. A LiDAR scan establishes 3D surface geometry and local thickness; the arm descends to apply controlled compression, and the distributed force map is processed to generate a 3D hardness surface and predict myopathy severity class.}
\label{fig:2}
\end{figure*}

\begin{figure*}[!t]
\centering
\includegraphics[width=\textwidth]{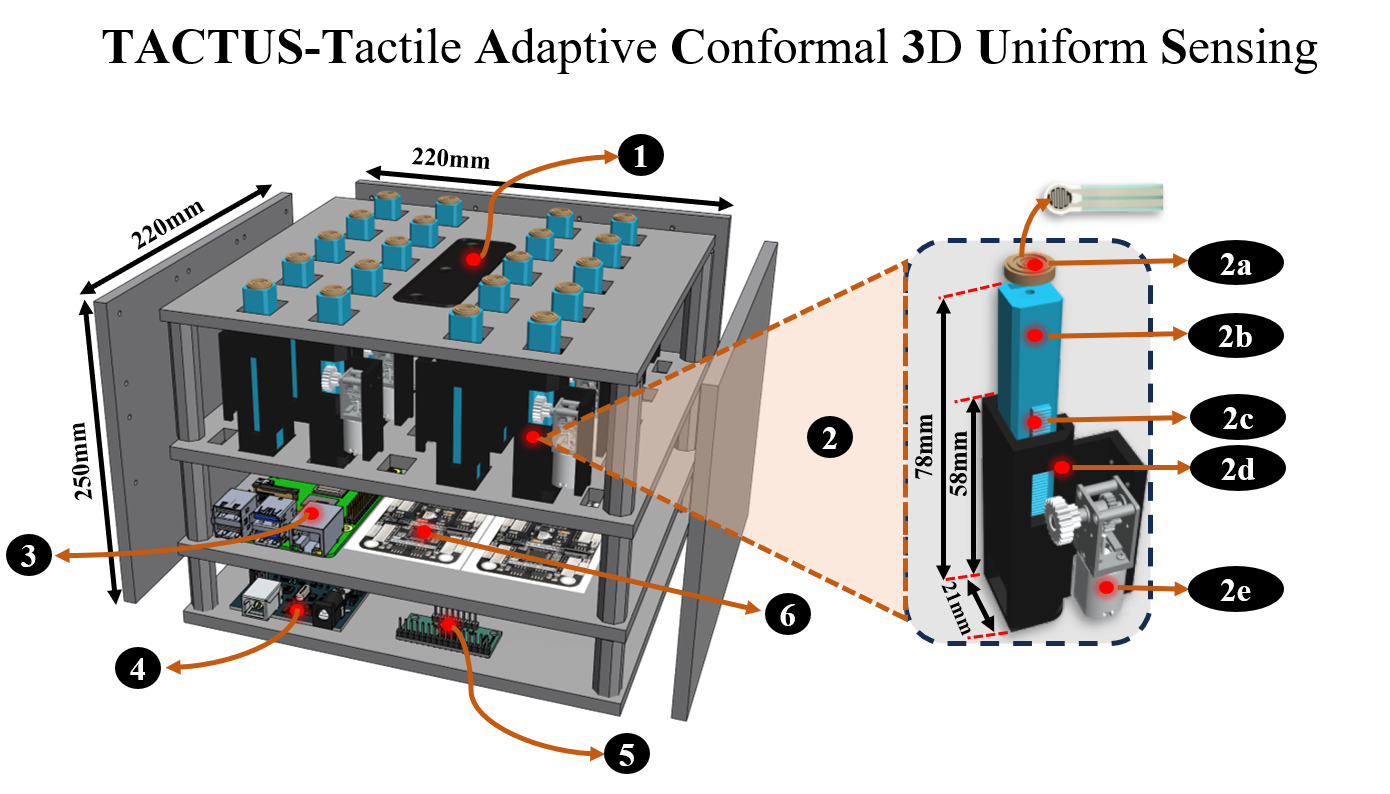}
\caption{Exploded hardware architecture of the TACTUS device: (1) Luxonis OAK-D Lite LiDAR/depth camera; (2) tactuator units each comprising (2a) probe tip with integrated force sensor, (2b) DFRobot SEN0299 thin-film piezoresistive force sensor, (2c) rack-and-pinion linear gear, (2d) ABS enclosure, and (2e) N20 quadrature encoder motor; (3) Raspberry Pi~4B compute module; (4) Arduino Mega microcontroller; (5) 16-channel analogue multiplexer (CD74HC4067); and (6) 4-Channel Encoder Motor Driver board with onboard voltage regulation circuit.}
\label{fig:3}
\end{figure*}

The TACTUS device is a modular, Polylactic Acid extrusion chassis (220~mm~(L)~$\times$~220~mm~(W)~$\times$~250~mm~(H)) designed to mount on the tool-flange of a Universal Robots UR5e collaborative arm. The uppermost tier houses the Luxonis OAK-D Lite depth camera. The central tier contains the 4$\times$5 array of 20 tactuator units at a 25~mm center-to-center pitch, spanning an active footprint of approximately 100~mm~$\times$~125~mm. The lower tier accommodates a Raspberry Pi 4B single-board computer (Raspberry Pi Ltd., Cambridge, United Kingdom), an Arduino Mega 2560 microcontroller (Arduino, Monza, Italy), five 4-Channel Encoder Motor Driver boards with onboard voltage-regulation circuitry (Hiwonder, Shenzhen, China), and a CD74HC4067 16-channel analog/digital multiplexer breakout board (LY170, Ximimark). Each motor-driver board independently controls four N20 tactuator motors, providing independent control of all 20 tactuators. The motor-driver boards interface with the Raspberry Pi through the I\textsuperscript{2}C communication network. A Raspberry Pi expansion board (Hiwonder, Shenzhen, China) provides convenient access to the SDA, SCL, 5~V, and GND connections for interfacing the I\textsuperscript{2}C hardware. The 20 force sensors described in Section~II-C are acquired independently through the Arduino Mega, with 15 sensors connected directly to its analog inputs and the remaining five routed through the CD74HC4067 multiplexer. The Raspberry Pi is powered by a dedicated 5~V DC adapter, the motor driver boards and tactuator motors are powered by a separate external 5~V DC supply, the Arduino Mega is powered through the Raspberry Pi, and a 12~V-powered cooling fan is mounted to prevent thermal throttling of the Raspberry Pi during operation. TACTUS attaches to the UR5e tool-flange through a custom 3D-printed mounting interface consisting of a cup-shaped socket fixed to the tool-flange and a mating solid cylindrical post fixed to the top of the TACTUS chassis, which couple together to rigidly locate the device on the arm. Total device mass is approximately 2.1~kg, within the UR5e end-effector payload capacity of 5~kg. A schematic overview of the complete robotic palpation workflow from LiDAR surface scan through conformal compression to 3D hardness map generation and myopathy classification is provided in Fig.~\ref{fig:2}, and the exploded hardware architecture of the device is shown in Fig.~\ref{fig:3}. The external envelope (220~$\times$~220~$\times$~250~mm) and the 4$\times$5 tactuator layout were dimensioned from the average morphometry of a commercial broiler breast fillet, approximately 168--184~mm in length, 90--93~mm in width, and 35--45~mm in cranial thickness \cite{ref26}, so that the array positions its 20 measurement points across a representative fillet, while the 40~mm tactuator travel (Section~II-C) matches the average maximum fillet thickness.

\subsubsection{Tactuator Design and Actuation}
Each of the 20 tactuator units integrates five sub-components within compact Polylactic Acid (PLA) housing (21~mm~$\times$~21~mm~$\times$~58~mm). Actuation was provided by a 6~V right-angle GA12-N20 DC gearmotor with an integrated dual-channel Hall effect quadrature encoder (generic OEM, China, purchased through eBay). The motor drives a custom 3D-printed rack-and-pinion linear gear assembly fabricated in PLA, yielding a theoretical linear displacement resolution of 0.08~mm per encoder count and a full vertical travel of 40~mm per tactuator. This travel range is sized to exceed the average maximum thickness of a commercial broiler breast fillet, ensuring that each probe can reach and compress the fillet surface across the full-size range while accommodating local surface-height variation of up to $\pm$20~mm relative to the mean surface plane. The square probe tip (1~$\times$~1~cm, 100~mm\textsuperscript{2} contact area, 3D-printed PLA) transmits a defined compressive load to the fillet surface. Each unit mounts via M3 fasteners to aluminum cross-rails in the central chassis tier. The principle of independently actuated pin arrays that conform to arbitrary surface geometry follows earlier shape-display systems such as inFORM \cite{ref27}, here adapted for instrumented force measurement rather than physical shape rendering.

\subsubsection{Force Sensing of Tactuator}
Force measurement at each tactuator is provided by a Gravity thin-film pressure sensor (Model SEN0616; DFRobot Co., Ltd., Shanghai, China) affixed onto the probe tip. SEN0616 is a compact analog thin-film force-sensitive resistor whose nanometre-scale pressure-sensitive layer forms conductive bridging contacts under applied load, reducing its electrical resistance as the applied force increases \cite{ref28}. Technical specifications relevant to the TACTUS application are summarised in Table~\ref{tab:sensor}. The sensor is bonded onto the 1~$\times$~1~cm square probe tip so that load is transmitted through the probe cross-section (contact area~=~100~mm\textsuperscript{2}), providing a well-defined and spatially consistent contact geometry. Sensor resistance is transduced as an analogue voltage through a voltage divider (10~k$\Omega$ fixed resistor, $V_{ref}$~=~5~V) and sampled by a 10-bit ADC channel on the Arduino Mega. An Arduino Mega 2560 equipped with a KEYESTUDIO MEGA Sensor Shield V1 was used for data acquisition. Fifteen thin-film pressure sensors (Sensors 1 to 15) were connected directly to the Arduino Mega 2560 analogue input pins A1 through A15. The remaining five sensors (Sensors 16 to 20) were interfaced through a CD74HC4067 16-channel analogue multiplexer, with the multiplexer signal output connected to analogue input A0. The required multiplexer channels were sequentially selected using four digital address lines to acquire measurements from Sensors 16 to 20. Each channel is sampled five times per acquisition with 2~ms inter-sample delay, and the median output is reported.

\begin{table*}[!t]
\caption{Technical Specifications of the DFRobot SEN0616 Thin-Film Piezoresistive Pressure Sensor as Deployed in Each TACTUS Tactuator Unit \cite{ref28}}
\label{tab:sensor}
\centering
\resizebox{\textwidth}{!}{%
\begin{tabular}{@{}p{3.2cm}p{7cm}p{7cm}@{}}
\toprule
\textbf{Parameter} & \textbf{Specification} & \textbf{Relevance to TACTUS} \\
\midrule
Model / SKU & DFRobot SEN0616 (Gravity thin-film pressure sensor) & Low-cost analog thin-film FSR \\
Sensing principle & Piezoresistive thin film; resistance decreases as applied force increases & Transduced to voltage, read by ADC \\
Measurement range & 30~g -- 1.5~kg ($\approx$0.29 -- 14.7~N) & Manufacturer range; empirically calibrated to fillet compression force (Section~II-E) \\
Overall dimensions & 57.5~$\times$~22~mm & Compact; bonded beneath the 1~$\times$~1~cm square probe tip \\
Sensor diameter \& thickness & 5~mm \& 0.4~mm & Ultra-low profile; mounts flush under probe tip \\
Operating voltage & 3.3 -- 5~V & Powered from the Arduino 5~V rail \\
Output signal & Analog (increases with applied force) & Sampled by 10-bit ADC (0--1023 counts) \\
Interface & Gravity PH2.0-3P (3-pin) & Plug-and-play wiring to the multiplexer \\
Weight & 2.9~g (sensor) & Negligible added mass per tactuator \\
Baseline resistance / hysteresis & Not specified by manufacturer; thin-film FSRs typically $>$10~M$\Omega$ unloaded with $\approx\pm$10\% hysteresis & Motivates empirical calibration (Section~II-E) and repeated-indentation averaging \\
\bottomrule
\end{tabular}%
}
\end{table*}

\subsubsection{LiDAR Surface Profiling and Thickness Normalization}
A stereo depth camera (OAK-D Lite; Luxonis Holding Corporation, Denver, CO, USA) is mounted 80~mm above the tactuator array plane. Prior to every compression test, TACTUS is positioned at a 450~mm standoff height above the fixture plate. Twenty depth frames were acquired and used to reconstruct a three-dimensional surface mesh, which was converted into a point cloud of the fillet surface within the 100~$\times$~125~mm measurement area. The point cloud was resampled onto a uniform 1~mm~$\times$~1~mm grid in the lateral (X-Y) plane to estimate local surface height and tissue thickness. Because the projected footprint of each 1~$\times$~1~cm tactuator hole on the fillet surface is known directly from the reconstructed mesh before any compression occurs, the contact location of every probe, together with the local fillet thickness at that location, is determined in advance of descent; the UR5e can additionally reposition TACTUS within the scan area to maximize the number of tactuators that achieve valid contact with the fillet. To aid consistent sample placement, a 22~$\times$~22~cm reference square is overlaid on the live camera view directly beneath the depth camera, indicating to the operator where the fillet should be positioned prior to scanning. For each tactuator location $i$, two geometric quantities are extracted from the reconstructed three-dimensional surface. The first is the local surface height, $\Delta z_i$, defined as the perpendicular distance between the dorsal (upper) surface of the fillet and the flat support plate on which the sample rests. This height is used to pre-position each tactuator, ensuring simultaneous contact with the irregular fillet surface. The second is the local fillet thickness, $T_i$, calculated as the average distance between the dorsal surface and the support plate within a 5~mm radius centered at the tactuator location. The estimated local thickness is subsequently used to set the target compression depth for each tactuator. The target compression depth for tactuator $i$ is defined as a fixed 20\% engineering strain relative to the local fillet thickness,
\begin{equation}
d_i = 0.20 \times T_i.
\label{eq:1}
\end{equation}
The compressive force recorded once this target depth is reached is calibrated to Newtons (Section~II-E) and used directly as the local force measurement at that tactuator location.

\subsubsection{UR5e Integration and Acquisition Protocol}
TACTUS is mounted on the UR5e tool-flange and controlled via the UR RTDE protocol at 500~Hz from a Raspberry Pi~4B running ROS~2 Humble on Ubuntu~22.04~LTS. A URDF model of the TACTUS device was created and integrated into this ROS~2 environment, and motion planning and control of the UR5e arm were performed using MoveIt. A standardized six-phase acquisition protocol is executed for each fillet. In phase~1, the fillet is placed on the fixture, and the arm moves TACTUS to the pre-scan position (450~mm standoff). In phase~2, the OAK-D Lite acquires 20 depth frames and reconstructs a three-dimensional surface mesh of the fillet; the Raspberry Pi computes $\Delta z_i$ and $T_i$ for all 20 points and transmits motor targets. In phase~3, all 20 motors simultaneously advance to $\Delta z_i - 2$~mm, bringing the array into conformal pre-contact configuration. In phase~4, the UR5e descends at 35~mm/s and all 20 channels are monitored; descent halts when the ADC value across channels exceeds 0, confirming surface contact. In phase~5, each motor independently advances by $d_i = 0.20 \times T_i$ at a constant speed of 2~mm/s. The compression speed was held constant for all measurements to ensure consistent loading conditions across all fillets. Although compression speed may influence the mechanical response of viscoelastic tissues, it was treated as a controlled experimental parameter in this study, allowing direct comparison of hardness measurements among samples. Force is recorded at 100~Hz for 2~s (200 force measurements per tactuator) at target compression depth. In phase~6, the UR5e retracts, motors return to the home position, and all 20 force-displacement traces together with the corresponding mesh, coordinate, and thickness data, are compiled and saved to a CSV file for each fillet. Total cycle time per fillet is approximately \textasciitilde30 seconds, measured from LiDAR-based mesh reconstruction at the home position through compression and return to the home position. As an inherent safety feature of the UR5e collaborative platform, motion is automatically halted via an emergency stop if the measured contact force at the tool flange exceeds a preset threshold at any point during the descent or compression sequence.

\subsection{Sensor Calibration Protocol}
\subsubsection{Sensing Principle and Rationale for Empirical Calibration}
The thin-film force sensor (Section~II-C) is a force-sensitive resistor (FSR) for which applied load forms conductive bridging contacts within a nanometre-scale pressure-sensitive layer, lowering the element resistance from an unloaded value exceeding 10~M$\Omega$ toward below 200~k$\Omega$ at full load \cite{ref28}. The manufacturer specifies no force transfer function; the reference firmware returns only the raw 10-bit analogue-to-digital (ADC) value, because the resistance-force behaviour of an FSR is intrinsically non-linear and depends strongly on contact area, backing hardness and the read-out divider, and is further degraded by approximately $\pm$10\% hysteresis and under 5\% static drift over 24~h (\cite{ref28,ref29}). A device- and probe-specific empirical calibration against a traceable force reference is therefore a standard practice for flexible resistive pressure sensors \cite{ref30}.

\subsubsection{Reference Instrumentation and Calibration Data Acquisition}
FSR sensor calibration was performed using a laboratory texture analyzer (MultiTest-dV motorized texture analyzer with VectorPro software v6.16.0.0; Mecmesin Ltd., Slinfold, West Sussex, United Kingdom), where the peak compression force required to achieve a fixed indentation depth of 10~mm served as the reference measurement. The force sensitive resistor (FSR) was bonded to the face of a 10~mm~$\times$~10~mm square compression probe (25~mm\textsuperscript{2}) such that the texture analyzer and the FSR simultaneously recorded the same compression event during each indentation, yielding directly paired force and sensor measurements. In addition, a 5~mm diameter circular probe (19.6~mm\textsuperscript{2}) was evaluated to establish the probe geometry conversion described in Section~II-E.

For each indentation, the FSR response was continuously recorded throughout the 0 to 10~mm compression stroke, producing a displacement sweep consisting of 150 samples. Each sample contained the raw analog-to-digital converter (ADC) value, the filtered ADC value, and the corresponding divider voltage calculated as
\begin{equation}
V = ADC \times 5/1023.
\label{eq:2}
\end{equation}
The maximum value of the filtered sensor response, corresponding to the instant of maximum indentation, was extracted and paired with the peak compression force measured by the texture analyzer using matching sample and measurement identifiers. Calibration data were collected from 12 chicken breast fillets, each evaluated at up to nine spatial locations, resulting in a total of 108 paired sensor force observations used for calibration and validation.

\subsection{Calibration Model Selection}
Candidate transfer functions (linear, proportional through-origin, quadratic and power-law) together with regularised and ensemble machine-learning models were compared under leave-one-sample-out cross-validation (LOSO-CV), in which all points of a given fillet are withheld and predicted from the remaining fillets. A simple linear model proved the most robust: additional polynomial or machine-learning complexity did not improve cross-validated accuracy, and the fitted power-law exponent (approximately 1.03) confirmed a near-proportional response. The adopted calibration, converting the peak raw ADC value (RawPeak~=~maximum Raw\_ADC recorded over the stroke) to peak compression force, is shown in Eq.~(\ref{eq:3}):
\begin{equation}
F_{peak}(N) = 0.0524 \times \mathrm{RawPeak} - 2.64,
\label{eq:3}
\end{equation}
with an equivalent through-origin form
\begin{equation}
F_{peak} = 0.0422 \times \mathrm{RawPeak},
\label{eq:4}
\end{equation}
and, expressed in voltage terms ($V_{peak} = 0.00489 \times \mathrm{RawPeak}$),
\begin{equation}
F_{peak} = 11.15 \times V_{peak} - 2.64.
\label{eq:5}
\end{equation}
The model explained $R^2 = 0.61$ of the variance (cross-validated $R^2 = 0.43$; root-mean-square error 5.1~N; mean absolute error 3.3~N). Calibrating to the average rather than the peak force produced a smoother and more accurate relationship ($F_{avg} = 0.0167 \times \mathrm{RawPeak} - 0.37$; cross-validated $R^2 = 0.58$). Because a single FSR reading is bounded by the sensor hysteresis and drift noted above, the mean of repeated indentations is reported for each location \cite{ref29}.

\subsubsection{Cross-Probe Force Conversion}
To relate forces obtained with the two probe geometries, per-sample mean forces were regressed, yielding
\begin{equation}
F_{circular}(N) = 0.50 \times F_{square} + 1.52 \quad (R^2 = 0.64),
\label{eq:6}
\end{equation}
or, in the robust simple form,
\begin{equation}
F_{circular} = 0.72 \times F_{square}.
\label{eq:7}
\end{equation}
Normalised by contact area this represents an approximately 3.5-fold higher contact stress under the smaller circular probe, consistent with indentation mechanics and confirming that force does not scale by the naive probe-area ratio.

\subsection{Statistical Analysis}
All statistical analyses were performed in Python (SciPy v1.11), with statistical significance assessed at $\alpha = 0.05$ unless otherwise noted. Differences in texture-analyzer and TACTUS force among woody breast severity categories, both at the point level and within each anatomical region (cranial, medial, caudal), were assessed using one-way and two-way analysis of variance (ANOVA), with pairwise comparisons performed where the overall category effect was significant. Agreement between texture-analyzer and TACTUS fillet-level median force was evaluated using Pearson correlation to test linear association and Spearman rank correlation to test monotonic association, together with linear regression and Bland-Altman analysis to quantify mean bias and 95\% limits of agreement. Spatial dependence of TACTUS force was assessed using semivariogram analysis (Section~II-F), in which the Spearman correlation between pairwise separation distance and semivariance was tested for the pooled dataset and separately within each severity category. Differences in local change entropy among categories (Section~II-F) were tested using the non-parametric Kruskal-Wallis test, because the underlying nearest-neighbor force differences were not normally distributed, with pairwise post hoc comparisons performed using Mann-Whitney~U tests and entropy evenness differences among categories additionally assessed using bootstrap resampling. Local heterogeneity was mapped descriptively across the sensing grid (Section~II-F) and interpreted alongside the number of contributing pair comparisons (N) as a measure of statistical support, rather than being subjected to a separate categorical hypothesis test.

\subsubsection{Semivariogram Analysis}
A semivariogram is a geostatistical method used to evaluate how the similarity of measurements changes as a function of spatial separation. It compares pairs of observations collected at different locations and calculates semivariance, which represents one half of the average squared difference between paired measurements within a given separation distance, or lag \cite{ref31,ref32}. Lower semivariance at shorter distances indicates that nearby locations tend to have more similar values, whereas increasing semivariance with distance indicates that measurements become progressively less similar as spatial separation increases. This relationship is commonly interpreted as evidence of spatial dependence or spatial autocorrelation in the measured property and is widely used to characterize spatial variability in agricultural and environmental data \cite{ref32,ref31,ref33}. In this study, the semivariance is calculated as
\begin{equation}
\hat{\gamma}(B_k) = \frac{1}{2N(B_k)} \sum_{(i,j) \in B_k} \left[Z(x_i) - Z(x_j)\right]^2,
\label{eq:8}
\end{equation}
\begin{equation}
B_k = \left\{ (i,j) : d_{ij} \in [h_k^-, h_k^+) \right\}.
\label{eq:8b}
\end{equation}
Here, $B_k$ denotes the $k$-th distance bin, $N(B_k)$ is the number of point pairs in that bin, $d_{ij}$ is the Euclidean distance between locations $x_i$ and $x_j$, and $Z(x_i)$ and $Z(x_j)$ are the measured values at those locations. The five distance bins are fixed, unequal-width intervals defined once from the pairwise Euclidean distances between TACTUS sensing-hole coordinates and reused for every semivariogram and short-distance analysis: [3.0, 4.5), [4.5, 6.5), [6.5, 8.5), [8.5, 10.5), and [10.5, 19) cm. Because the sensing locations sit on a fixed 4-row~$\times$~5-column grid rather than being randomly scattered, pairwise distances do not spread continuously but cluster into a small number of discrete geometric tiers set by the grid spacing (adjacent holes, one-step-diagonal pairs, two-step pairs, and so on, with a long tail of far-apart corner-to-corner pairs). The bin edges were placed in the gaps between these natural clusters, rather than as equal-width or quantile bins, so that each bin isolates one geometric tier of hole-to-hole spacing instead of splitting a cluster of near-identical distances across two bins; this is also why the final bin (10.5--19~cm) is substantially wider, as it pools the sparser, more scattered long-distance pairs. The first bin (3.0--4.5~cm) corresponds to the nearest-neighbor ``adjacent holes'' tier used elsewhere in the local change entropy and local heterogeneity analyses (Section~II-F).

\subsubsection{Local Change Entropy Analysis}
Local change entropy was used as a descriptive measure of the magnitude and distribution of short distance force changes between neighboring TACTUS sensing locations. For each nearest neighbor pair, the absolute force difference was calculated as
\begin{equation}
|\Delta F_{ij}| = |F_i - F_j|,
\label{eq:9}
\end{equation}
where $F_i$ and $F_j$ represent the force measurements at neighboring sensing locations $i$ and $j$. The resulting absolute force differences were grouped into six bins, and the probability associated with each bin was calculated as
\begin{equation}
p_k = \frac{n_k}{N},
\label{eq:10}
\end{equation}
where $n_k$ is the number of nearest neighbor differences falling within bin $k$, and $N$ is the total number of nearest neighbor pairs within the corresponding category. Shannon entropy was then calculated from the binned distribution of local force differences as
\begin{equation}
H = -\sum_{k=1}^{K} p_k \log_2(p_k),
\label{eq:11}
\end{equation}
where $K$ represents the total number of bins and $p_k$ is the probability of observations occurring in bin $k$. Shannon entropy provides a measure of the uncertainty or diversity of a probability distribution, with higher values indicating that observations are distributed across a wider range of bins and lower values indicating stronger concentration within a smaller number of bins \cite{ref34}. To facilitate comparison among categories, Shannon entropy was normalized by the theoretical maximum entropy for the number of bins used. The maximum entropy for $K$ equally populated bins is $H_{\max} = \log_2(K)$, and normalized Shannon entropy, also referred to here as entropy evenness, was calculated as
\begin{equation}
E = \frac{H}{H_{\max}} = \frac{H}{\log_2(K)},
\label{eq:12}
\end{equation}
where $E$ ranges from 0 to 1. Values approaching 1 indicate that local force differences are distributed relatively evenly across the available bins, whereas lower values indicate that the differences are concentrated within a smaller number of magnitude ranges. This normalization follows the general principle of dividing observed Shannon entropy by its maximum possible value, as commonly used in evenness analysis \cite{ref35}.

\subsubsection{Local Heterogeneity Mapping}
Local spatial heterogeneity was evaluated using only measurements collected by the TACTUS device to determine how much the measured force changed between neighboring sensing locations on the same fillet. The TACTUS sensing surface consisted of 20 fixed sensing positions arranged in a 4~row~$\times$~5~column grid. Each sensing position was assigned to a hole number from 1 to 20, which was converted to its corresponding grid location as
\begin{equation}
r_i = \left\lfloor \frac{h_i - 1}{5} \right\rfloor + 1,
\label{eq:13}
\end{equation}
\begin{equation}
c_i = \left( (h_i - 1) \bmod 5 \right) + 1,
\label{eq:14}
\end{equation}
where $h_i$ is the TACTUS hole number, $r_i$ is its grid row, and $c_i$ is its grid column. Importantly, the grid represents the physical arrangement of the TACTUS sensing locations and was not generated from the 3D fillet geometry. Row indices use integer floor division, and column indices use the modulo operation to recover the within-row position.

For each fillet, only active TACTUS sensing locations that were in contact with the fillet and produced a valid force measurement were included. Inactive locations were excluded rather than being assigned a force of zero. Therefore, if only 7 of the 20 sensing locations were active for a particular fillet, all calculations for that fillet were performed exclusively using those 7 measured locations. Furthermore, sensing locations were compared only with other active locations from the same fillet. Measurements from different fillets were never directly paired.

For every pair of active sensing locations within a fillet, the planar separation distance was calculated as
\begin{equation}
d_{ij} = \sqrt{(x_i - x_j)^2 + (y_i - y_j)^2},
\label{eq:15}
\end{equation}
where $d_{ij}$ is the distance between sensing locations $i$ and $j$ in cm, and $(x_i,y_i)$ and $(x_j,y_j)$ are their respective planar coordinates. Local heterogeneity was specifically evaluated using pairs separated by 3.0 to 4.5~cm, representing nearby or approximately neighboring sensing locations. Thus, only pairs satisfying $3.0 \le d_{ij} \le 4.5$~cm were retained for the local heterogeneity analysis.

The force difference for each qualifying pair was calculated directly from the TACTUS force measurements as $\Delta F_{ij} = |F_i - F_j|$, where $F_i$ and $F_j$ are the TACTUS force measurements in N at active sensing locations $i$ and $j$, respectively, and $\Delta F_{ij}$ represents the absolute difference in force between the two neighboring locations. The absolute difference was used because the objective was to quantify the magnitude of local force variation, regardless of which location had the greater force. Each qualifying pair contributed its $\Delta F_{ij}$ value to both sensing locations involved in the comparison. After these calculations were performed separately within every fillet, the resulting neighboring-pair differences were pooled across all fillets belonging to the same woody breast severity category. The local heterogeneity associated with grid position $g$ was then calculated as
\begin{equation}
H_g = \frac{1}{n_g} \sum_{p=1}^{n_g} \Delta F_{g,p},
\label{eq:16}
\end{equation}
where $H_g$ is the mean local force heterogeneity at grid position $g$, $\Delta F_{g,p}$ is the absolute TACTUS force difference from the $p$-th qualifying neighboring pair involving that position, and $n_g$ is the total number of qualifying neighboring-pair comparisons contributing to that grid position across all fillets within the category.

The value $n$ (individual measurement points pooled across the whole fillet) in each cell represents the total number of valid short-distance pair comparisons contributing to that cell, pooled across all fillets within the corresponding category, rather than the number of active sensors or fillets. Thus, cells with larger $n$ have greater measurement support, while low-$n$ cells should be interpreted cautiously. The four maps were generated separately for Normal, Mild, Moderate, and Severe fillets, with larger force-difference values indicating greater local mechanical heterogeneity between neighboring active sensing locations. The caudal, medial, and cranial labels provide approximate anatomical orientation only and were not used in calculating local heterogeneity.

\section{Results and Discussion}

\subsection{Comparative Assessment of Fillet Hardness Using Texture Analyzer and TACTUS}

\begin{table}[!t]
\caption{Texture-Analyzer Force Measurement by Category, Point-Level N, Mean $\pm$ SD, Median, and 95\% CI (N)}
\label{tab:table3}
\centering
\begin{tabular}{@{}lccccc@{}}
\toprule
\textbf{Class} & \textbf{n} & \textbf{Mean} & \textbf{SD} & \textbf{Median} & \textbf{95\% CI} \\
\midrule
Normal   & 179 & 4.25 & 4.75 & 2.75 & [2.88, 4.61] \\
Mild     & 315 & 4.40 & 4.54 & 3.30 & [3.37, 4.61] \\
Moderate & 180 & 7.18 & 5.54 & 5.35 & [5.60, 7.56] \\
Severe   & 225 & 7.57 & 5.85 & 5.80 & [6.03, 8.06] \\
\bottomrule
\end{tabular}
\end{table}

\begin{figure}[!t]
\centering
\includegraphics[width=\columnwidth]{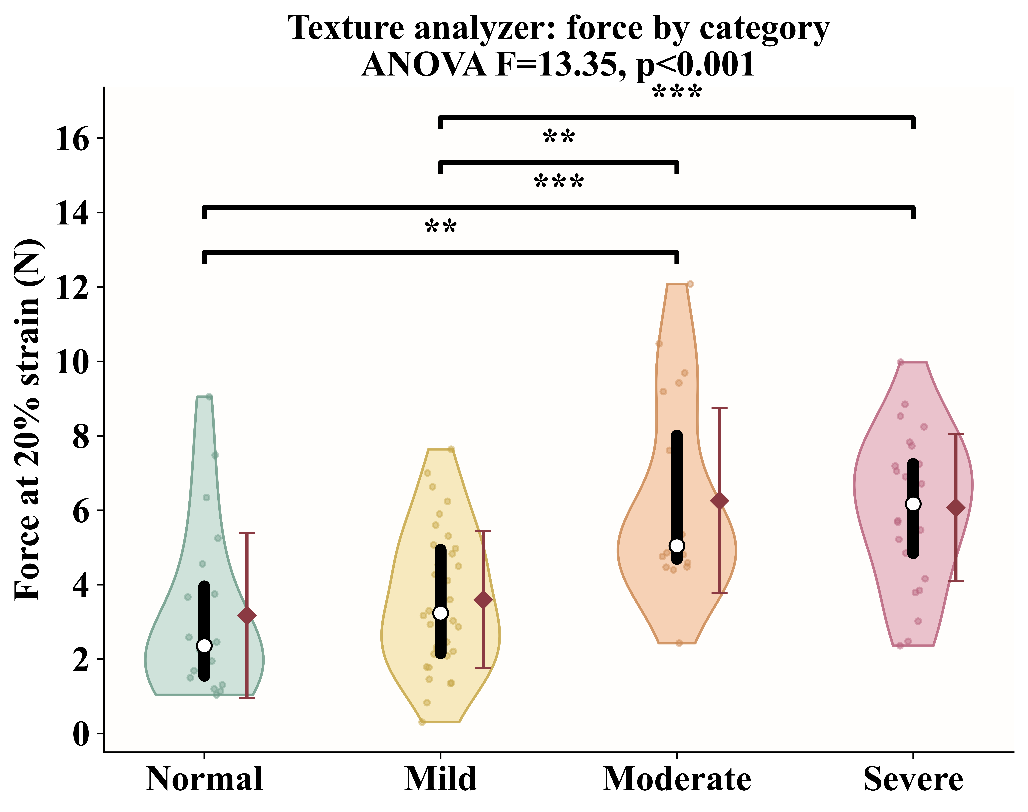}
\caption{Texture-analyzer force measurement by category.}
\label{fig:4}
\end{figure}

\begin{figure*}[!t]
\centering
\includegraphics[width=\textwidth]{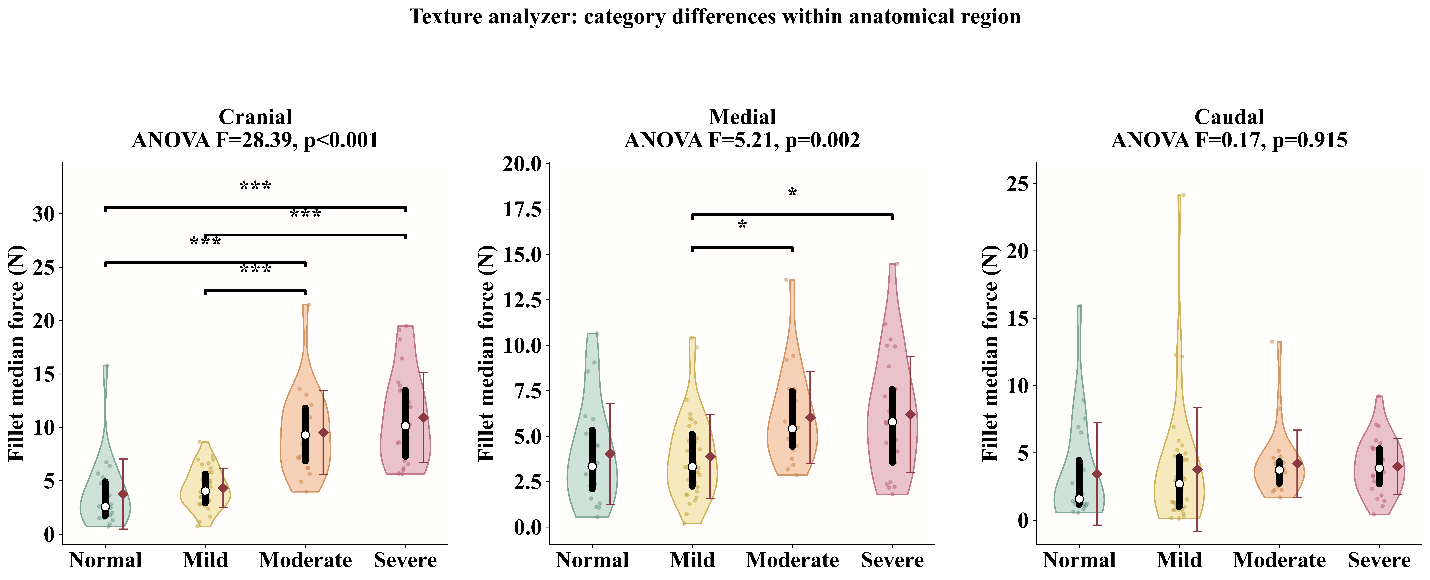}
\caption{Texture-analyzer force measurement differences across categories by different fillet regions.}
\label{fig:5}
\end{figure*}

The texture-analyzer measurements showed a clear increase in force with increasing severity category (Table~\ref{tab:table3} and Fig.~\ref{fig:4}). At the point level, mean force increased from 4.25~N in Normal fillets to 4.40~N in Mild, 7.18~N in Moderate, and 7.57~N in Severe fillets, with corresponding median values of 2.75, 3.30, 5.35, and 5.80~N, respectively (Table~\ref{tab:table3}). The overall effect of category was highly significant (ANOVA, $F = 13.35$, $p < 0.001$), and the pairwise comparisons shown in Fig.~\ref{fig:4} indicate significant differences primarily between the lower severity groups and the Moderate and Severe categories. However, the magnitude of these category differences varied considerably by anatomical region (Fig.~\ref{fig:5}). A strong category effect was observed in the cranial region ($F = 28.39$, $p < 0.001$), where force increased markedly from Normal and Mild to Moderate and Severe fillets. A smaller but still significant category effect was observed in the medial region ($F = 5.21$, $p = 0.002$), whereas no significant category-related difference was detected in the caudal region ($F = 0.17$, $p = 0.915$). Collectively, Table~\ref{tab:table3} and Figs.~\ref{fig:4} and~\ref{fig:5} demonstrate that the increased force associated with greater severity is spatially dependent, with the strongest differentiation among severity categories occurring in the cranial region, followed by the medial region, while the caudal region provides little discrimination among categories.

\begin{table}[!t]
\caption{TACTUS Force Measurement by Category, Point-Level N, Mean $\pm$ SD, Median, and 95\% CI (N)}
\label{tab:table4}
\centering
\begin{tabular}{@{}lccccc@{}}
\toprule
\textbf{Class} & \textbf{n} & \textbf{Mean} & \textbf{SD} & \textbf{Median} & \textbf{95\% CI} \\
\midrule
Normal   & 200 & 5.30 & 3.22 & 4.48 & [4.35, 5.55] \\
Mild     & 413 & 5.16 & 3.43 & 4.37 & [4.44, 5.23] \\
Moderate & 308 & 6.94 & 3.65 & 7.28 & [6.72, 8.28] \\
Severe   & 382 & 6.67 & 3.55 & 7.01 & [6.84, 8.34] \\
\bottomrule
\end{tabular}
\end{table}

\begin{figure}[!t]
\centering
\includegraphics[width=\columnwidth]{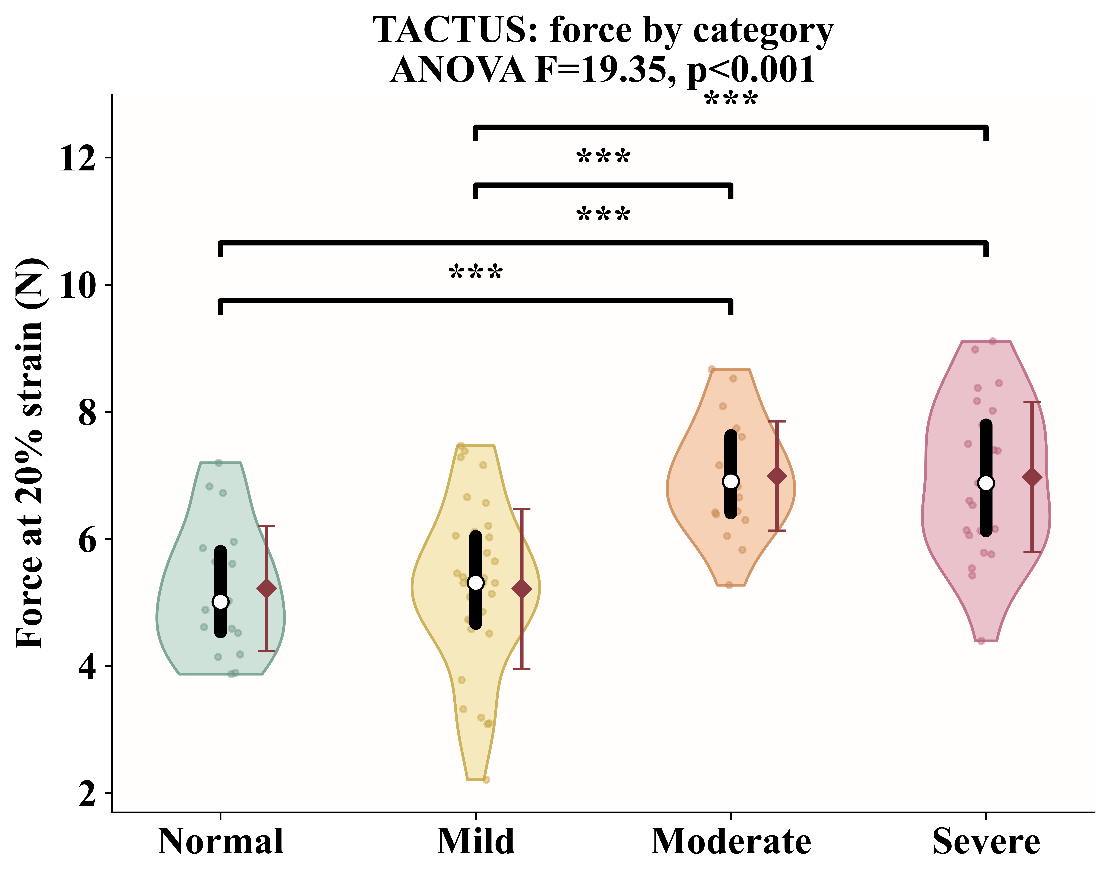}
\caption{Distribution of TACTUS force measurements across woody breast severity categories.}
\label{fig:6}
\end{figure}

\begin{figure*}[!t]
\centering
\includegraphics[width=\textwidth]{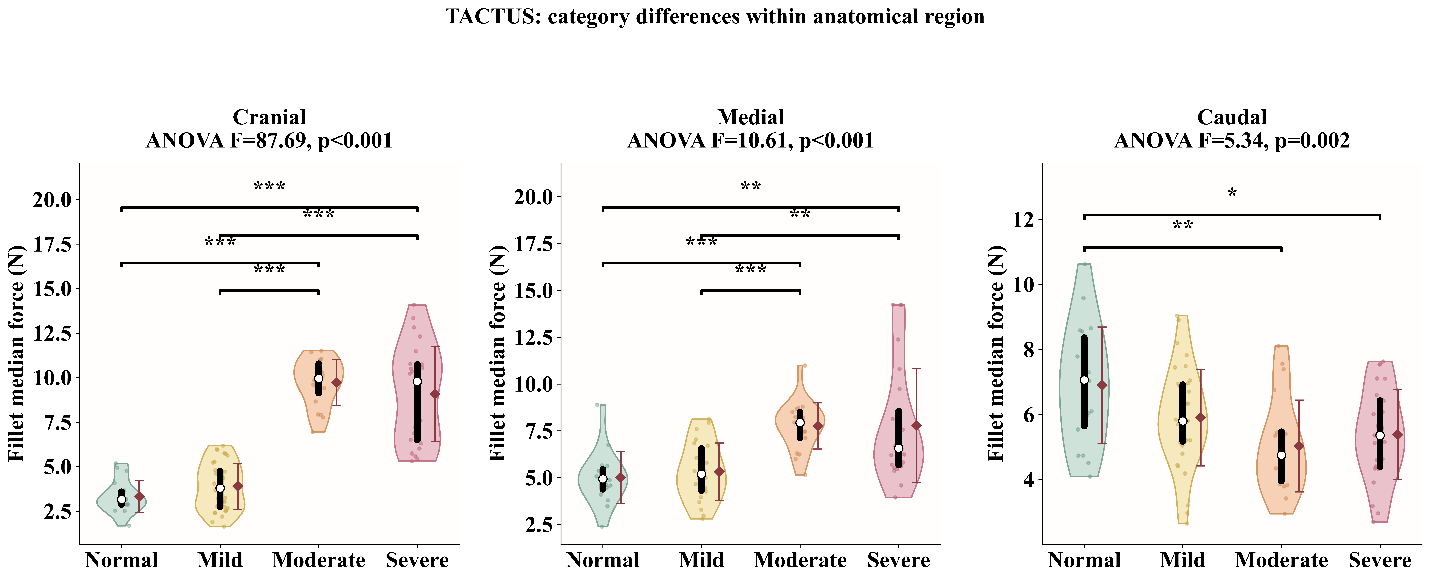}
\caption{Distribution of TACTUS force measurements across woody breast categories within anatomical regions.}
\label{fig:7}
\end{figure*}

TACTUS measurements showed significant differences among severity categories (Table~\ref{tab:table4} and Fig.~\ref{fig:6}), with higher forces generally observed in Moderate and Severe fillets than in Normal and Mild fillets (ANOVA, $F = 19.35$, $p < 0.001$). Region-specific analysis showed the strongest category effect in the cranial region ($F = 87.69$, $p < 0.001$), followed by the medial region ($F = 10.61$, $p < 0.001$), while a smaller but significant effect was observed in the caudal region ($F = 5.34$, $p = 0.002$) (Fig.~\ref{fig:7}).

Overall, the texture analyzer and TACTUS showed similar severity-related and spatial patterns, with the greatest differentiation occurring in the cranial region and reduced differentiation toward the caudal region. However, TACTUS showed stronger category effects in the cranial and medial regions and detected a significant caudal effect that was not observed with the texture analyzer ($F = 0.17$, $p = 0.915$; testing the null hypothesis of no difference in mean texture-analyzer force among the four WB severity categories within the caudal region; the non-significant result indicates the texture analyzer did not detect category-related differences in that region), suggesting greater differentiation among severity categories across anatomical regions.

\subsection{Fillet-Level Comparison and Agreement Between Texture Analyzer and TACTUS}
Instrument agreement was assessed by correlating the fillet-level median forces obtained from the texture analyzer and TACTUS for fillets measured using both devices. Pearson correlation tested the linear association between measurements, while Spearman correlation assessed the monotonic rank association, providing a direct evaluation of concordance between the two instruments.

\begin{figure}[!t]
\centering
\includegraphics[width=\columnwidth]{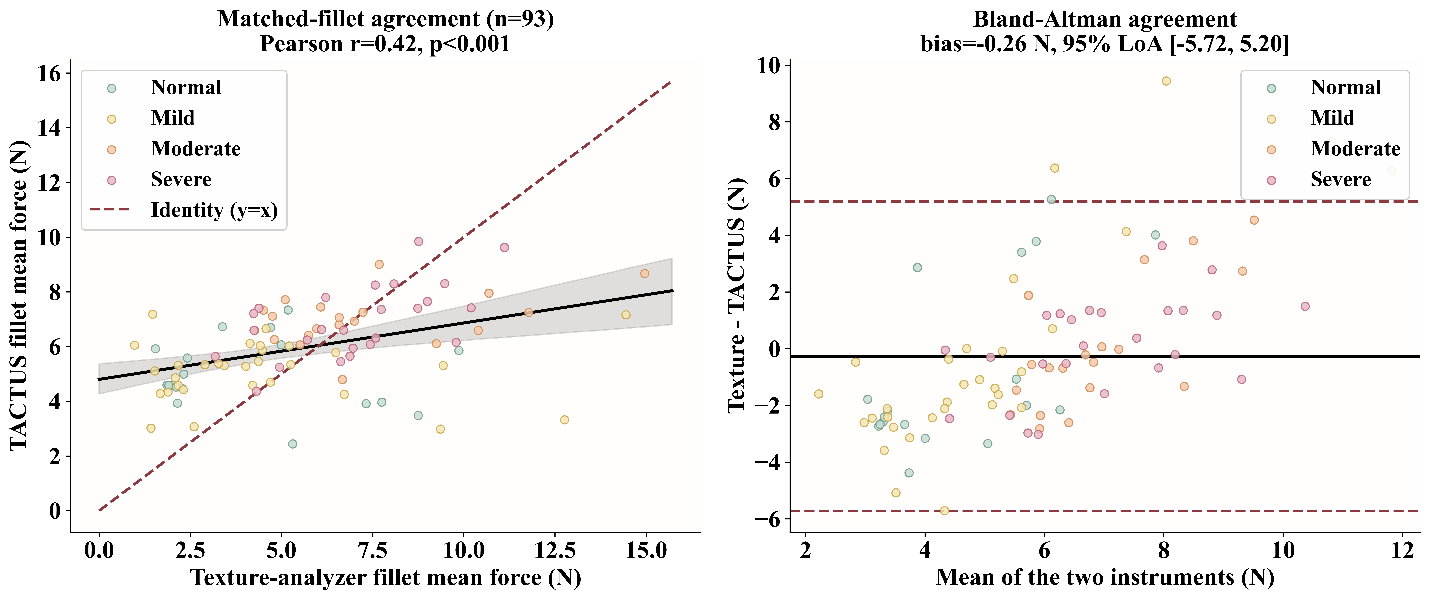}
\caption{Matched-fillet correlation and Bland-Altman agreement between texture analyzer and TACTUS.}
\label{fig:8}
\end{figure}

Across all the fillets measured by both instruments, TACTUS measurements showed a significant moderate positive association with texture analyzer measurements (Pearson $r = 0.441$, 95\% CI [0.262, 0.591], $p < 0.001$; Spearman $\rho = 0.457$, $p < 0.001$), indicating that fillets with greater hardness measured by the texture analyzer generally also produced higher TACTUS measurements. Linear regression further supported this relationship ($R^2 = 0.195$), although the modest $R^2$ and slope below 1 indicate that the two instruments did not produce equivalent absolute force measurements. Bland-Altman analysis showed a mean bias of $-0.26$~N (Texture minus TACTUS), with 95\% limits of agreement ranging from $-5.72$ to 5.2~N, demonstrating considerable variation between measurements. Importantly, the two instruments measured different locations on the same fillet rather than identical points. Because Woody Breast hardness is spatially heterogeneous, the observed differences may reflect both instrument-related variation and local tissue variability. Therefore, these findings support fillet-level agreement in relative hardness trends but should not be interpreted as direct point-to-point agreement or evidence that TACTUS and the texture analyzer are interchangeable for absolute force measurement.

\subsection{Spatial Dependence of TACTUS Force Using Semivariogram Analysis}
Spatial dependence of TACTUS force was assessed using semivariogram analysis, as described in Section~II-F.

\begin{table}[!t]
\caption{Spatial Correlation Between Distance and TACTUS Force Semivariance by Woody Breast Category}
\label{tab:table5}
\centering
\begin{tabular}{@{}lcc@{}}
\toprule
\textbf{Category} & \textbf{n pairs\textsuperscript{a}} & \textbf{Spearman $\rho$} \\
\midrule
Normal   & 605  & 0.060 \\
Mild     & 1229 & 0.135$^{***}$ \\
Moderate & 1114 & 0.100$^{**}$ \\
Severe   & 1503 & 0.166$^{***}$ \\
\bottomrule
\end{tabular}
\\[4pt]
\raggedright\footnotesize Superscripts indicate significance of each category's own distance-vs-$|\Delta F|$ correlation against the null of $\rho=0$ ($^{*}p<0.05$, $^{**}p<0.01$, $^{***}p<0.001$); Normal's correlation is not statistically significant ($p=0.137$) and \textsuperscript{a}(n~pairs) denotes the number of point-pair comparisons contributing to a given estimate, and is distinct from the point-level sample size $n$ reported in Tables~\ref{tab:table3} and~\ref{tab:table4}.
\end{table}

\begin{figure}[!t]
\centering
\includegraphics[width=\columnwidth]{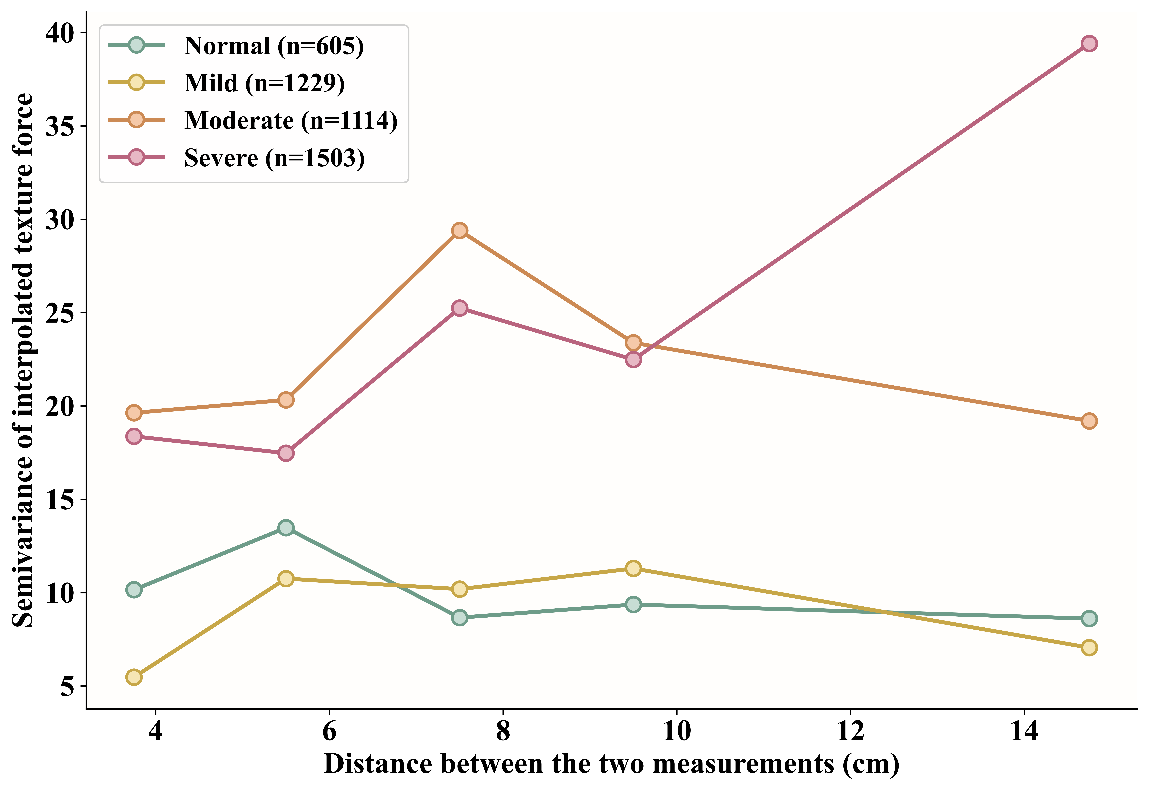}
\caption{TACTUS force semivariance as a function of distance across woody breast categories.}
\label{fig:9}
\end{figure}

For the TACTUS measurements, the empirical semivariograms showed generally higher semivariance in Moderate and Severe fillets than in Normal and Mild fillets, indicating greater spatial heterogeneity in force measurements as shown in Fig.~\ref{fig:9}. The pooled Spearman correlation between separation distance and semivariance was $\rho = 0.112$, $p < 0.001$, $n = 4{,}451$ pairs, demonstrating a weak but statistically significant positive spatial relationship (Table~\ref{tab:table5}). When examined by woody breast severity category, the same general trend was observed. Severe fillets showed the strongest spatial association with $\rho = 0.166$, followed by Mild with $\rho = 0.135$, Moderate with $\rho = 0.100$, and Normal with $\rho = 0.060$. Significant associations were observed for Mild, Moderate, and Severe fillets, while the Normal category was not significant, likely because Normal fillets lack the spatially organized fibrotic hardening that produces a distance-dependent force gradient, so nearby and distant sensing locations are not measurably more or less similar than one another. Overall, these results indicate that TACTUS force measurements exhibit a modest spatial structure across the fillet surface, with nearby sensing locations tending to have more similar force values than locations separated by larger distances. The slightly stronger relationship observed in the Severe category may suggest that advanced wooden breast produces more spatially organized regions of increased hardness, whereas Normal fillets exhibit comparatively less structured spatial variation. This interpretation is consistent with Parajuli et al. \cite{ref6}, who used systematic grid-based compression-force mapping and reported that the fillet area classified as woody-breast-affected increased from approximately 16\% in Normal fillets to 35\% in Moderate and 50\% in Severe fillets, with affected tissue forming increasingly large, spatially contiguous regions rather than scattered isolated points as severity increased, and with the histopathological progression described by Sihvo et al. \cite{ref8}, in which early, focal myodegeneration coalesces into more diffuse, spatially extensive fibrosis at advanced disease stages. Both findings support greater spatial organization of hardened tissue in Severe fillets relative to Normal. However, the correlation magnitudes remain small, so the results indicate weak spatial dependence rather than a strong spatial gradient. Consistent with this pattern, the single highest semivariance value observed across all categories and distance bins in Fig.~\ref{fig:9} occurs in the Severe category's largest-distance bin (10.5--19~cm); because TACTUS's 100~$\times$~125~mm sensing footprint is small, pairs separated by this maximum distance predominantly compare cranial-region to caudal-region locations, and Severe fillets showed the largest cranial-versus-caudal disparity among the four categories (cranial $F = 87.69$ vs. caudal $F = 5.34$; Section~III-A), so this bin is expected to capture the largest force contrast of any category-distance combination.

\subsection{Local Change Entropy of Nearest Neighbor TACTUS Force Differences}
Local change entropy of nearest-neighbor TACTUS force differences was used to characterize the magnitude and distribution of short-range force variation across the fillet, as described in Section~II-F.

\begin{table}[!t]
\caption{Local Change Entropy and Nearest Neighbor TACTUS Force Differences Across Woody Breast Categories}
\label{tab:table6}
\centering
\resizebox{\columnwidth}{!}{%
\begin{tabular}{@{}lcccc@{}}
\toprule
\textbf{Category} & \textbf{Bins} & \textbf{Median $|\Delta F|$} & \textbf{Raw entropy (bits)} & \textbf{Evenness (0--1)} \\
\midrule
Normal   & 6 & 1.70 & 2.38$^{a}$  & 0.92$^{a}$  \\
Mild     & 6 & 1.01 & 1.95$^{b}$  & 0.76$^{b}$  \\
Moderate & 6 & 2.61 & 2.29$^{ac}$ & 0.89$^{ac}$ \\
Severe   & 6 & 2.71 & 2.52$^{c}$  & 0.98$^{c}$  \\
\bottomrule
\end{tabular}%
}
\end{table}

Higher median $|\Delta F|$ indicates larger typical force differences between neighboring sensing locations. Shannon entropy describes the diversity of those local force differences across the six magnitude ranges, while entropy evenness describes how uniformly the differences are distributed among those ranges, with values closer to 1 indicating a more even distribution. Importantly, these metrics do \textbf{not} directly measure Woody Breast severity, fibrosis, or any specific pathological change. They quantify only the spatial heterogeneity of the measured force response. That heterogeneity can arise from several factors, including local thickness variation, anatomical location, tissue structure, and myopathy-related changes, and these effects were not separated in this analysis. Therefore, a strictly increasing pattern from Normal to Severe should not necessarily be expected. Mild fillets showed the lowest median $|\Delta F|$ (1.01~N), entropy (1.95~bits), and evenness (0.76), indicating relatively small and concentrated local force differences. Normal fillets had a higher entropy (2.38~bits) and evenness (0.92) despite being unaffected by Woody Breast, which may reflect natural anatomical and thickness-related variation, particularly differences between thicker and thinner regions of the fillet. Moderate and Severe fillets showed larger median neighboring force differences of 2.61 and 2.71~N, respectively, with Severe also showing the highest entropy (2.52~bits) and evenness (0.98). Thus, the results indicate differences in local mechanical heterogeneity, rather than a direct severity scale, and the observed variation may represent a combination of normal geometry and myopathy-related tissue changes.

A significant overall difference in $|\Delta F|$ was observed among categories ($H = 72.64$, $p < 0.001$). Pairwise Mann-Whitney~U tests showed that Normal differed significantly from Mild ($p = 0.002$), Moderate ($p = 0.019$), and Severe ($p = 0.035$), while Mild also differed from Moderate and Severe (both $p < 0.001$). Moderate and Severe were not significantly different ($p = 0.874$). Bootstrap comparisons of entropy evenness further showed that Mild had significantly lower evenness than Normal and Moderate, while Severe had significantly higher evenness than Normal, Mild, and Moderate. Overall, these results confirm significant differences in local force heterogeneity among categories, with Mild showing the most homogeneous local response and Severe showing the greatest variability in neighboring force differences.

\subsection{Local Heterogeneity Mapping of TACTUS Force}
Local spatial heterogeneity of TACTUS force was mapped across the sensing grid, as described in Section~II-F.

\begin{figure}[!t]
\centering
\includegraphics[width=\columnwidth]{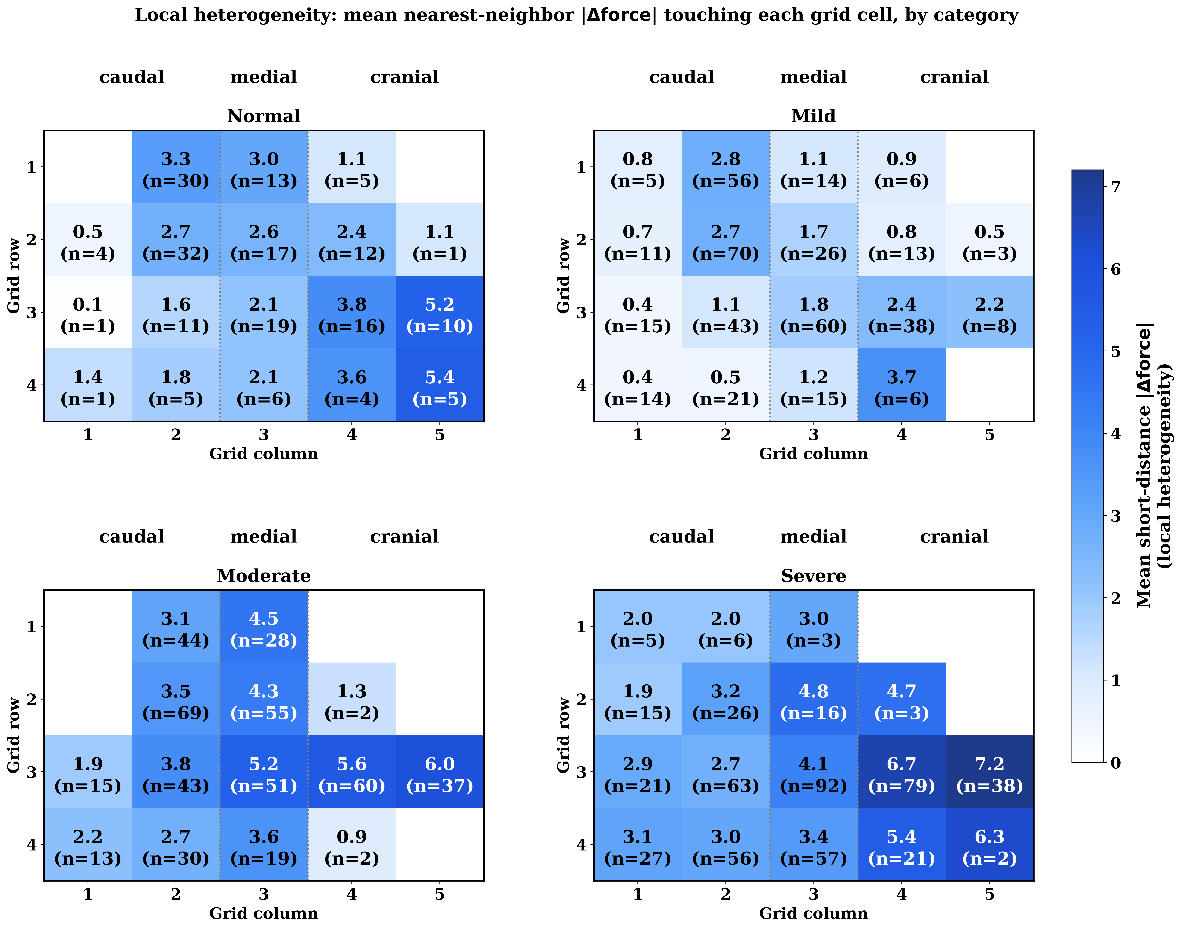}
\caption{Local TACTUS force heterogeneity based on mean nearest-neighbor absolute force differences ($|\Delta F|$) across grid positions by woody breast category.}
\label{fig:10}
\end{figure}

The local heterogeneity maps showed generally lower neighboring force differences in Normal and Mild fillets, while Moderate and Severe fillets exhibited greater heterogeneity, particularly toward the cranial region as shown in Fig.~\ref{fig:10}. Severe fillets showed the strongest local differences, reaching approximately 6 to 7~N in some locations, suggesting greater short-range variation in hardness. However, cells with low $n$ should be interpreted cautiously because they are supported by fewer pair comparisons. Unlike the semivariogram, which evaluates how force dissimilarity changes with distance across the grid, the local heterogeneity map identifies where the largest short-distance force changes occur. The entropy analysis provides another complementary measure by describing the overall diversity and distribution of short-distance force differences within each category, but it does not retain their spatial locations. Thus, the semivariogram describes distance-dependent heterogeneity, entropy describes the diversity of local force differences, and the local heterogeneity map describes where those local differences occur.

\subsection{Discussion}
The spatial patterns captured by TACTUS are consistent with, and substantially extend, the limited prior evidence that woody breast (WB) hardness is not a scalar property but a spatially organized phenotype. Both instruments used in this study showed the strongest categorical differentiation in the cranial region and progressively weaker differentiation toward the caudal region (Fig.~\ref{fig:5}, Fig.~\ref{fig:7}), reproducing the cranial predominance of fibrotic hardening reported by Parajuli et al. \cite{ref6}, who applied a systematic grid-based compression force (CF) protocol at 10 to 38 discrete locations per fillet using 6, 13, and 25~mm probes to map the heterogeneous spatial distribution of WB hardness. TACTUS reproduced this core finding while compressing the acquisition time from the 15 to 20 minutes per fillet required for sequential manual grid palpation \cite{ref6} to approximately 30 seconds through simultaneous, LiDAR-guided, motorized actuation of all 20 tactuators. TACTUS also detected a statistically significant effect in the caudal region, suggesting that dense, simultaneous, perpendicular-contact sampling may capture subtle regional differences that sequential single-probe measurements may miss due to greater sensitivity to variations in probe placement and contact angle. This pattern also aligns quantitatively with Parajuli et al. \cite{ref6}, who reported that WB percentage predicted from 25~mm-probe compression-force heatmaps increased from approximately 16\% in Normal fillets to 35\% in Moderate fillets and 50\% in Severe fillets, a monotonic severity gradient that mirrors the significant categorical increase observed here in both texture-analyzer (median 2.75 to 5.80~N) and TACTUS force values (Tables~\ref{tab:table3} and~\ref{tab:table4}), despite the two studies using entirely different force-sensing hardware and acquisition protocols.

The comparatively modest fillet-level agreement between TACTUS and the texture analyzer (Pearson $r = 0.441$, $R^2 = 0.195$) and the weak, though significant, semivariogram spatial correlations (pooled $\rho = 0.112$, ranging from 0.060 in Normal to 0.166 in Severe fillets) are plausibly explained in part by probe-geometry effects rather than by measurement error alone. The cross-probe conversion established in Section~II-E showed that force does not scale linearly with contact area between the two probe geometries tested (1~cm and 5~mm), corresponding to an approximately 3.5-fold difference in contact stress once normalized by area. This is consistent with Parajuli et al.'s \cite{ref6} finding that smaller-diameter probes (6~mm) were more sensitive to spatially localized fibrotic foci than larger probes (25~mm), because larger contact areas mechanically average the compressive response over a larger, and often mechanically heterogeneous, tissue volume. TACTUS's 1~$\times$~1~cm (100~mm\textsuperscript{2}) square tactuator tips, while enabling full-fillet simultaneous coverage, likely perform similar spatial averaging at each of the 20 measurement points, which would tend to smooth sharp, sub-centimeter hardness gradients and could partially account for the moderate, rather than strong, correlation observed against single-point texture-analyzer measurements taken at a different, non-identical location on the same fillet.

An important limitation of TACTUS is that it provides a depth-normalized surface hardness map rather than a volumetric measure of hardness through the tissue. This distinction is relevant because WB hardness can vary both spatially and with tissue depth. Sun et al. \cite{ref40} reported greater compression force in the cranial region, decreasing toward the caudal region, consistent with the regional pattern observed with TACTUS (cranial $F = 74.21$, medial $F = 13.13$, caudal $F = 3.22$). They also reported greater hardness in superficial than internal tissue, suggesting that TACTUS's surface-referenced compression may capture the tissue layer where WB-associated hardening is more pronounced, although it cannot directly resolve hardness variation through tissue depth. More recently, Li et al. \cite{ref41} used B-mode ultrasound and deep learning to stratify internal WB regions, identifying pathological boundaries at approximately 55\%, 62\%, and 65\% relative depth for Mild, Moderate, and Severe fillets, respectively, with classification accuracies up to 98.71\%. Together, these findings highlight that WB heterogeneity occurs both across the fillet surface and through tissue depth. While TACTUS characterizes spatial surface hardness, its current single-compression-depth protocol cannot resolve internal hardness or fibrosis gradients, suggesting that depth-resolved measurements could further strengthen WB characterization.

The local change entropy results offer a plausible mechanistic window into WB progression when interpreted alongside the established histopathology of the condition. Mild fillets exhibited the lowest local force-difference entropy (the most homogeneous neighboring-pair response), consistent with early-stage WB manifesting as small, discrete, internally embedded stiff foci within a surrounding matrix of otherwise near-normal soft tissue \cite{ref7}, such that most neighboring sensor pairs sample predominantly the surrounding normal tissue and register small, uniform force differences. Severe fillets, by contrast, showed significantly higher entropy evenness, indicating that neighboring-pair force differences were spread broadly across the full range of magnitudes. This is consistent with the progression from focal myodegeneration to diffuse, coalescing fibrosis with regenerative activity described histologically by Sihvo et al. \cite{ref8} and Soglia et al. \cite{ref5}, in which regions of dense fibrotic plaque become interspersed with residual less-affected tissue at increasingly fine spatial scales, producing local force gradients of highly variable magnitude across the fillet.

Recent optical imaging methods have also achieved strong WB classification performance, including 93\% accuracy with SIRI and surface profilometry \cite{ref36}, 86.34\% with hyperspectral scattering imaging \cite{ref37}, and strong prediction of MORS texture features using Vis-NIR hyperspectral imaging \cite{ref38}. However, these approaches relied on categorical or localized mechanical reference measurements rather than spatially resolved hardness ground truth. Pairing such imaging methods with spatially registered TACTUS hardness maps could therefore provide richer ground truth and improve the spatial interpretability of future WB detection models.

\subsection{Limitations and Future Direction}
Several limitations of TACTUS should be considered. The DFRobot SEN0616 thin-film force sensor is nonlinear and susceptible to hysteresis and drift \cite{ref29,ref28}, resulting in moderate calibration performance ($R^2 = 0.61$; LOSO-CV $R^2 = 0.43$; RMSE~=~5.1~N). This sensor uncertainty may partly explain the moderate agreement with the texture analyzer ($R^2 = 0.195$; Bland-Altman limits: $-5.52$ to 3.78~N), indicating that TACTUS is currently more appropriate for relative spatial hardness characterization than precise absolute force measurement.

Second, the fixed 4~$\times$~5 array with 25~mm spacing and 1~$\times$~1~cm tactuator tips limits spatial resolution and may miss small localized fibrotic regions, particularly in Mild WB, while the larger contact area may spatially average local hardness variations \cite{ref7,ref6}. Third, validation was limited to a single genetic line, production system, and raw-meat condition without direct histological validation. Spatial results should also be interpreted cautiously where the number of contributing measurement pairs were low. Fourth, the current TACTUS unit is a laboratory-grade, first-generation prototype rather than an industrial-grade system: its electronics and PLA housings are not sealed or waterproofed, and only a single physical unit has been built and evaluated to date, with an improved version planned for future development. The device has completed approximately 200 to 300 compression cycles without mechanical failure, providing preliminary evidence of short-term durability, although long-term reliability under high-volume, washdown-capable industrial conditions remains to be established.

Building on these limitations, future TACTUS development will prioritize multimodal fusion with hyperspectral and transillumination imaging to achieve pixel-level WB characterization beyond the current 25~mm grid resolution. Spatially registered TACTUS hardness maps could provide ground truth for co-registered optical images, addressing the limitation of training spatially rich imaging data using scalar or single-point labels. HSI could capture spectral changes associated with WB, while transillumination may provide complementary structural contrast from fibrotic tissue. Previous studies combining optical imaging with surface profiling support the feasibility of this approach (\cite{ref36,ref39}). Future work should also investigate whether transillumination-derived optical density can provide information about internal lesion depth, which surface compression and reflectance HSI cannot directly resolve \cite{ref40,ref41}. The second priority is redesigning TACTUS with a higher-density array of smaller probes, which may improve detection of localized fibrotic regions by reducing spatial averaging \cite{ref6}. Replacing the current thin-film FSRs with miniature load cells could also reduce nonlinearity, hysteresis, and drift, thereby improving absolute force accuracy and cross-instrument agreement. However, increasing probe density and incorporating load cells will require redesigning the tactuator housing, actuation system, and acquisition electronics while balancing spatial resolution, device size, cost, and measurement time.

\section{Conclusion}
This study introduced TACTUS, a tactile array for automated robotic palpation, thickness-normalized, three-dimensional hardness mapping of broiler breast fillets across 20 spatially distributed sensing locations. TACTUS identified clear regional and severity-related differences in WB hardness, including significant effects across cranial, medial, and caudal regions. Spatial analyses using semivariograms, local change entropy, and local heterogeneity mapping further demonstrated that hardness variation is spatially structured and localized across the fillet rather than uniformly distributed. Moderate fillet-level agreement with the texture analyzer indicated that TACTUS reliably captured relative hardness trends, although absolute measurements differed due to factors such as sampling location and probe geometry.

Finally, TACTUS provides an automated and operator-independent platform for generating spatially registered mechanical hardness maps of WB fillets. The system enables simultaneous characterization of hardness magnitude, regional variation, and local spatial heterogeneity, providing information that cannot be obtained from conventional single-point measurements. These capabilities establish TACTUS as a promising platform for spatial characterization of WB and for generating spatially resolved mechanical ground truth for future imaging-based detection and mapping systems.

\section*{Acknowledgments}
The authors acknowledge grant support by the University of Arkansas Experimental Station and the University of Arkansas College of Engineering, USDA National Institute of Food and Agriculture (No: 2023-70442-39232, 2024-67022-42882).

\end{document}